\documentclass[conference]{IEEEtran}
\usepackage{times}
\usepackage{array,multirow}
\usepackage[square,numbers]{natbib}
\usepackage{notoccite}
\usepackage{multicol}
\pdfoutput=1
\usepackage{graphicx}
\usepackage{tabularx}
\usepackage{subcaption}
\usepackage{multirow,multicol, array}

\begin{document}

\title{End-to-End Navigation in Unknown Environments using Neural Networks}

\author{Arbaaz Khan, Clark Zhang, Nikolay Atanasov, Konstantinos Karydis, Daniel D. Lee, Vijay Kumar}



%

\maketitle

\begin{abstract}
We investigate how a neural network can learn perception actions loops for navigation in unknown environments. Specifically, we consider how to learn to navigate in environments populated with cull-de-sacs that represent convex local minima that the robot could fall into instead of finding a set of feasible actions that take it to the goal. Traditional methods rely on maintaining a global map to solve the problem of over coming a long cul-de-sac. However, due to errors induced from local and global drift, it is highly challenging to maintain such a map for long periods of time. One way to mitigate this problem is by using learning techniques that do not rely on hand engineered map representations and instead output appropriate control policies directly from their sensory input. We first demonstrate that such a problem cannot be solved directly by deep reinforcement learning due to the sparse reward structure of the environment. Further, we demonstrate that deep supervised learning also cannot be used directly to solve this problem. We then investigate network models that offer a combination of reinforcement learning and supervised learning and highlight the significance of adding fully differentiable memory units to such networks. We evaluate our networks on their ability to generalize to new environments and show that adding memory to such networks offers huge jumps in performance. \\

\end{abstract}

\IEEEpeerreviewmaketitle

\section{Introduction}
Designing autonomous perception-action loops for robot navigation remains a challenging problem. In real world scenarios, the robot has access to limited information about its environment. Traditional methods such as Simultaneous Localization and Mapping (SLAM) solve this problem by creating a representation of the environment \cite{slam0}. Inside this representation, control commands are generated by solving for the most optimal path. Due to accumulated errors along the trajectory, the reconstructed map tends to be inconsistent with the ground truth map. This problem is compounded in large scale unstructured environments \cite{1slam}. The advent of neural networks in the past few years provide an interesting alternative to solve this problem \cite{2nnslam1},\cite{3nnslam2}. However, most of these works rely on incorporating the neural network into some part of the SLAM pipeline. It has been shown that training perception and control system end to end performs better than individually training each component \cite{4vmpol}. Policy search methods rely on learning from an expert. Thus, for the case of robot navigation, given an expert policy, a neural network can be trained to directly map sensor information to control commands. 

In this work we investigate the effectiveness of solving this problem with deep reinforcement learning and deep supervised learning. Specifically, we investigate a recently proposed approach for approximating value iteration with convolutional networks, termed value iteration networks (VIN) \cite{5vin}. In the original VIN paper, the evaluation is done by presenting the full map to the robot. However for most robot navigation problems, the environment is often unknown. Our experiments indicate that directly using VINs for navigating complex obstacles such as cul-de-sacs, fails when the only input is the immediate sensor information. We then investigate the impact of adding memory to these networks and show that adding memory drastically improves performance and the network learns the concept of cul-de-sac by generalizing to longer lengths not seen in the training set.

\section{Background}

Consider a bounded connected set $\mathcal{X}$ representing the workspace of a robot. Let $\mathcal{X}^{obs}$ and $\mathcal{X}^{goal}$, called the obstacle region and the goal region, respectively, be subsets of $\mathcal{X}$. Denote the obstacle-free portion of the workspace as $\mathcal{X}^{free} := \mathcal{X}\backslash\mathcal{X}^{obs}$. The dynamics of the robot are specified by the Probability Density Function (PDF) $p_f(\cdot \mid x_t, u_t)$ of the robot state $x_{t+1} \in \mathcal{X}$ at time $t+1$ given the previous state $x_t \in \mathcal{X}$ and control input $u_t\in \mathcal{U}$. We assume that the control input space $\mathcal{U}$ is a finite discrete set.\footnote{For instance, the control space $\mathcal{U}$ for a differential-drive robot in $SE(2)$ can be a set of motion primitives, parameterized by linear velocity, angular velocity and duration. For a quadrotor, $\mathcal{U}$ may be a set of short-range dynamically feasible motions.} 
The robot perceives its environment through observations $z_t \in \mathcal{Z}$ generated from a depth sensor (e.g., lidar, depth camera), whose model is specified by the PDF $p_h(\cdot \mid \mathcal{X}^{obs},x_t)$. The information available to the robot at time $t$ to compute the control input $u_t$ is $i_t := (x_{0:t},z_{0:t},u_{0:t-1},\mathcal{X}^{goal}) \in \mathcal{I}$, consisting of current and previous observations $z_{0:t}$, current and previous states $x_{0:t}$ and previous control inputs $u_{0:t-1}$.\\
\textbf{Problem.} Given an initial state $x_0 \in \mathcal{X}^{free}$ and a goal region $\mathcal{X}^{goal} \subset \mathcal{X}^{free}$, find a function $\mu: \mathcal{I} \rightarrow \mathcal{U}$, if one exists, such that applying the control $u_t := \mu(i_t)$ results in a sequence of states that satisfies $\{x_0,x_1,\ldots,x_T\} \subset \mathcal{X}^{free}$ and $x_T \in \mathcal{X}^{goal}$.

\begin{figure}
  \centering
  \includegraphics[width=\linewidth]{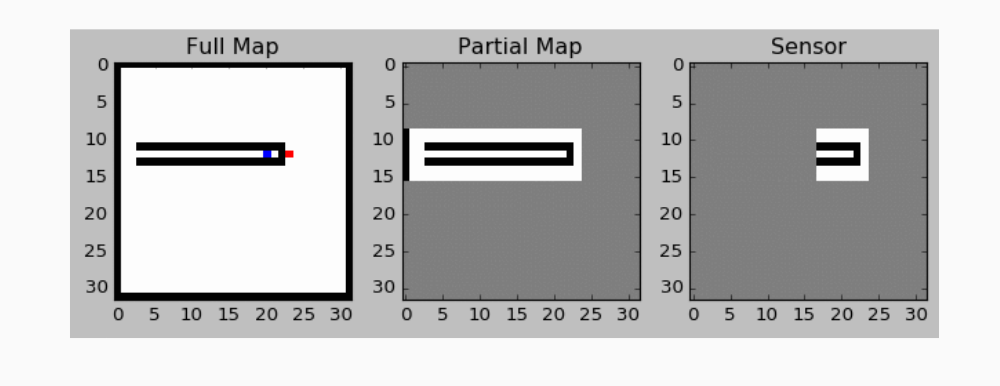}
  \caption{Simulated cul-de-sacs. The robot only sees the sensor input. Partial map represents the stitched images the robot has seen so far. In the full map the red dot represents the goal and the blue dot represents the robot}
\end{figure}

In the rest of the paper we consider a 2-D grid world, an instance of the feasible planning problem in which $\mathcal{X}$ is restricted to two dimensions and  $\mathcal{U}$:=${down,right,up,left}$. Planning needs sequential decision making. To investigate the ability of networks to react to sequential tasks we consider the case of cul-de-sacs as shown in Figure 1. We learn a feasible policy by using the outputs of an $A^*$ path planner for supervision \cite{6astar}. 

\subsection{Simulator}

For our task, we construct a simulation environment as seen in Fig 1. The robot at any point observes its immediate environment through the sensor input. We feed this patch of sensory information to our networks along with a reward prior that encodes information about the goal. We also construct a partial map by stitching together all the images the robot has seen till the current timestep. Every time the robot observes some new part of the environment, it is added to the partial map. Ideally, we want our robot to learn a representation of this partial map so that it can backtrack the cul-de-sac when it meets a dead end. The full map represents a top down view of the entire map where the blue dot represents the robot and the red dot represents the goal. The partial map represents a stitched together version of all states explored by the robot. The sensor input is the data available to the robot.

\section{Simulation and Results}

Solving this problem with deep reinforcement learning is extremely hard due to the sparsity of rewards, i.e the robot only gets a reward when it reaches the goal. We implement the DQN  architecture used in \cite{7dqn} to successfully play Atari games. However, due to the sparse reward structure of the environment, even after training for 1 million iterations the robot does not converge to the optimal policy. Using a supervised approach with a fully convolutional network does not work either. This is because some of the inputs from the sensor are mapped to two different control actions. For example, the sensor input while going into the cul-de-sac is the same when exiting the cul-de-sac. Value iteration networks offer some promise of solving this with their inherent ability to solve sequential tasks. Unfortunately, these do not work when given only the sensor input. An interesting observation is that when the value iteration networks are provided with the partial map instead of the sensor input alone, they perform very well and are able to work on lengths of up to 500 units when trained on lengths of 20 units. This indicates that there is a need for memory in our task since the partial map represents past state memory being fed into the network by the user. Since we are pursuing an end to end approach, we would like to have our network learn this map representation on its own.\\

We go back and try adding a LSTM to the convolutional network architecture tried above. We observe that this model performs poorly and is only able to get to the goal in some cases. Further, the model trained on small cul-de-sacs does not generalize to longer cul-de-sacs. On longer cul-de-sacs the robot does not explore the cul-de-sac all the way to the end. Instead, it turns around at a distance of 20 units implying that it has learned to turn around at a fixed distance of 20 units and hasn't learned the structure of the cul-de-sac. This turn around at a fixed distance behavior is consistent when the CNN+LSTM model is trained on cul-de-sacs of different lengths. Our next experiment involves adding a LSTM layer in the VIN architecture. We observe that the LSTM layer performs best when added after the attention module described in the VIN paper \cite{5vin}. With 256 hidden states, the LSTM is able to navigate cul-de-sacs of length upto 200 units when only being trained on lengths of 20 units. 
\begin{table}[t!]
	\vspace*{-0.2cm}
	\begin{center}
    \setlength\extrarowheight{4pt}
	\resizebox{0.48\textwidth}{!}
	{\begin{tabular}{||c|c|c||}
	\hline
	\textbf{\textit{Model}} & \textbf{\textit{Success ($\%$)}} & \textbf{\textit{Maximum generalization length}}  \\  \hline
	{DQN}  & 0 & 0 \\ \hline
	{CNN}  & 0 & 0 \\ \hline
	{CNN + LSTM}  & 40 & 20 \\ \hline
	{VIN}  & 0 & 0 \\ \hline
	{VIN + Partial Map}  & 100 & 500 \\ \hline
	{VIN + LSTM}  & \textbf{100} & \textbf{190} \\ \hline
	\end{tabular}}
	\end{center}
\caption{\textbf{\textit{RESULTS}}: All models are trained on cul-de-sac's of length 20 units. The success percentage represents the number of times the robot reaches the goal position in the test set. Maximum generalization length is the length of the longest cul-de-sac that the robot is able to successfully navigate after being trained on cul-de-sacs of length 20 units.}
\label{table:Results}
\vspace*{-0.2in}
\end{table}

\begin{figure}[h]
 
\begin{subfigure}{\linewidth}
\includegraphics[scale = 0.185]{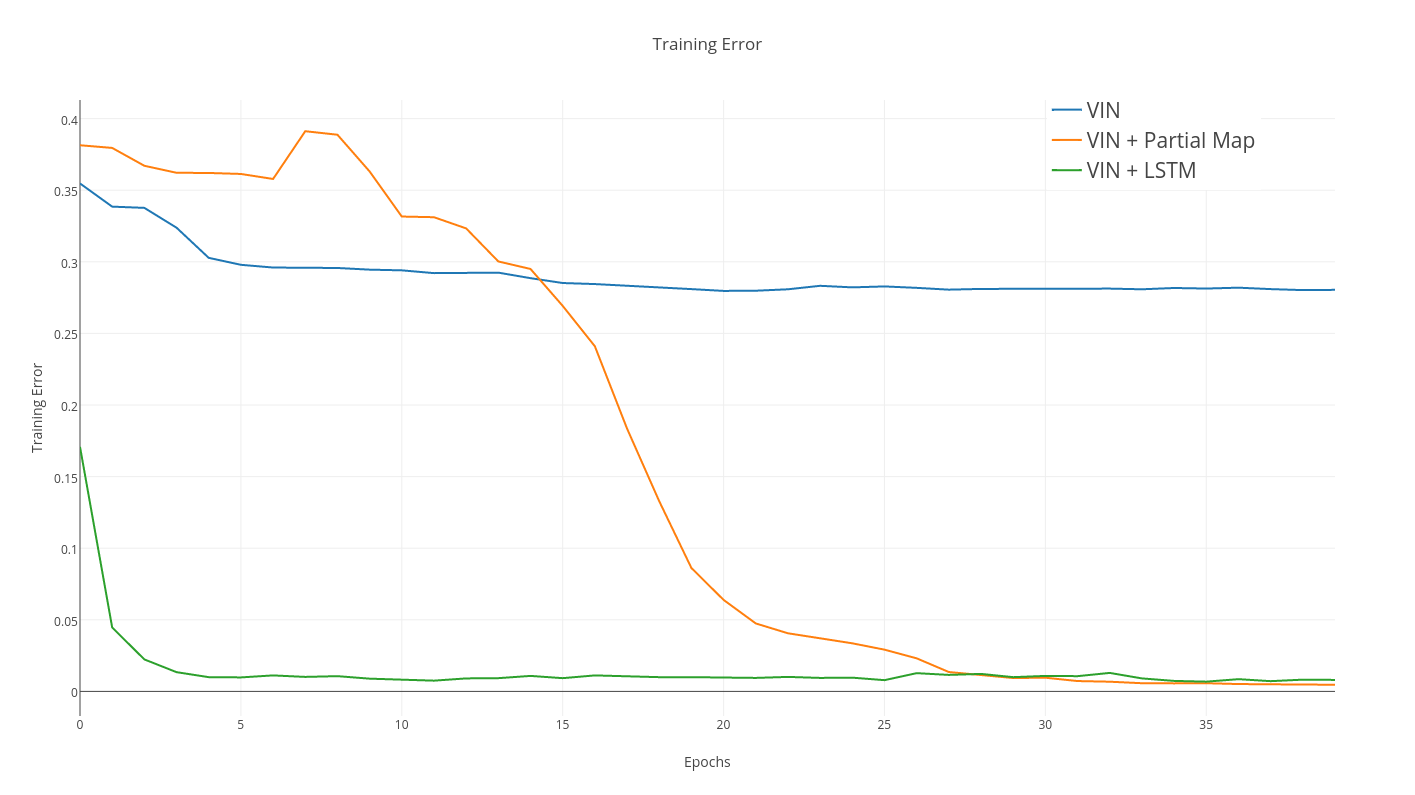} 
\caption{Training Error}
\label{fig:subim1}
\end{subfigure}
\begin{subfigure}{\linewidth}
\includegraphics[scale= 0.185]{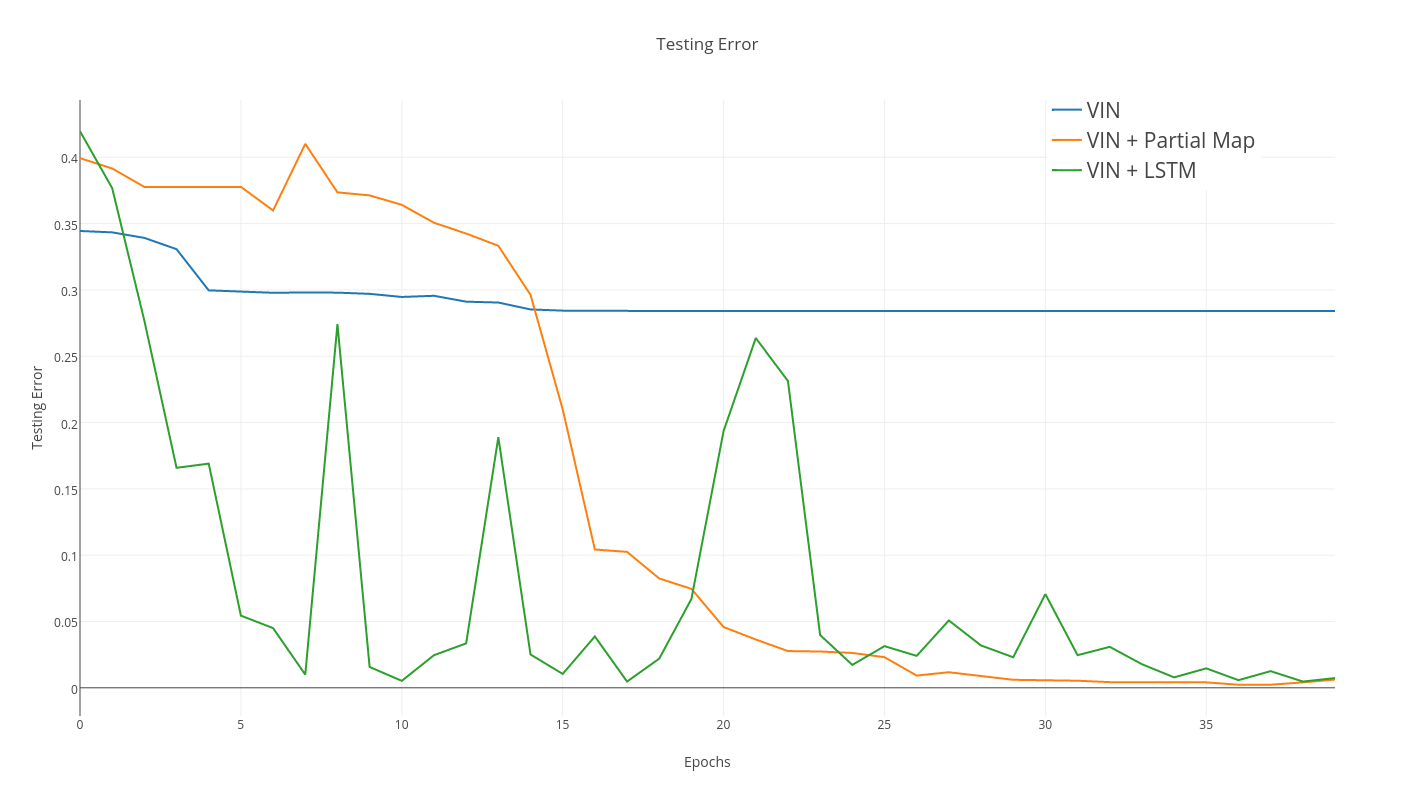}
\caption{Test Error}
\label{fig:subim2}
\end{subfigure}
 
\caption{The VIN + Partial Map is the case where a hand engineered map is presented to the network. Ideally, we want our networks to do as well as this model when presented with only the sensor input.}
\label{fig:image2}
\end{figure}

\section{Future Work and Conclusion}
The value iteration networks augmented with LSTMs perform poorly when trained on cul-de-sac’s of different orientations. Further, they are limited by the number of  This represents the need for a smarter memory structure. Our initial experiments with the Differentiable Neural Computer \cite{8dnc} have offered better results than just the LSTM. This indicates the significance of a smarter memory architecture geared specifically towards solving robot navigation tasks. In our future work, we would like to extend this work from simulation to a real robot. The input to the network would then be a lidar scan. The model can still be trained in simulation by modelling lidar noise and motion noise. Thus, in this work we have highlighted how an end to end navigation model can be used for goal driven navigation even in the presence of convex local minima. Another avenue to pursue would be to extend this planning scheme to 3 dimensional environments. Three dimensional environments present the robot with a lot of information not all of which is necessary for planning. In such a scenario learning to differentiate between what is redundant and what is not would be interesting. 
\label{sec:conclusion}

\bibliographystyle{IEEEtran}
\bibliography{references}

\end{document}